\documentclass[10pt,conference]{IEEEtran}
\IEEEoverridecommandlockouts
\usepackage{cite}
\usepackage{amsmath,amssymb,amsfonts,mathtools,bm}
\usepackage{amsthm}
\usepackage{algorithm}
\usepackage{algorithmic}
\usepackage{graphicx}
\usepackage{textcomp}
\usepackage{xcolor,float}
\usepackage{tabularx}
\usepackage{diagbox}
\usepackage{svg}
\usepackage{url} 
\usepackage{booktabs}
\usepackage{subfigure}
\usepackage{multirow}

\usepackage{booktabs}
\usepackage{pifont}
\usepackage{adjustbox}
\usepackage{subcaption}
\usepackage{subfig}
\usepackage{caption}

\allowdisplaybreaks
\renewcommand{\baselinestretch}{1}

\begin{document}

\title{UrbanMIMOMap: A Ray-Traced MIMO CSI Dataset with Precoding-Aware Maps and Benchmarks}

\author{
\IEEEauthorblockN{
Honggang Jia,
Xiucheng Wang,
Nan Cheng,
Ruijin Sun,
Changle Li
}
\IEEEauthorblockA{
School of Telecommunications Engineering, Xidian University, Xi'an, 710071, China\\
Email: jiahg@stu.xidian.edu.cn, 
xcwang\_1@stu.xidian.edu.cn,
dr.nan.cheng@ieee.org, \\
sunruijin@xidian.edu.cn,
clli@mail.xidian.edu.cn
}
}

    \maketitle

\IEEEdisplaynontitleabstractindextext

\IEEEpeerreviewmaketitle

\begin{abstract}
Sixth generation (6G) systems require environment-aware communication, driven by native artificial intelligence (AI) and integrated sensing and communication (ISAC). Radio maps (RMs), providing spatially continuous channel information, are key enablers. However, generating high-fidelity RM ground truth via electromagnetic (EM) simulations is computationally intensive, motivating machine learning (ML)-based RM construction. The effectiveness of these data-driven methods depends on large-scale, high-quality training data. Current public datasets often focus on single-input single-output (SISO) and limited information, such as path loss, which is insufficient for advanced multi-input multi-output (MIMO) systems requiring detailed channel state information (CSI). To address this gap, this paper presents UrbanMIMOMap, a novel large-scale urban MIMO CSI dataset generated using high-precision ray tracing. UrbanMIMOMap offers comprehensive complex CSI matrices across a dense spatial grid, going beyond traditional path loss data. This rich CSI is vital for constructing high-fidelity RMs and serves as a fundamental resource for data-driven RM generation, including deep learning. We demonstrate the dataset's utility through baseline performance evaluations of representative ML methods for RM construction. This work provides a crucial dataset and reference for research in high-precision RM generation, MIMO spatial performance, and ML for 6G environment awareness. The code and data for this work are available at: \url{https://github.com/UNIC-Lab/UrbanMIMOMap}.
\end{abstract}


\section{Introduction}
In sixth generation (6G) mobile communication systems, the rapid development of massive multiple-input multiple-output (Massive MIMO) and ultra-dense networks challenges channel state information (CSI) acquisition efficiency \cite{bjornson2017massive, sun2025acomprehesive}. Particularly in ultra-massive MIMO systems, the overhead of traditional pilot-based methods becomes prohibitive, which potentially reduces spectral efficiency and adaptation speed \cite{wang2024tutorial}. Concurrently, future 6G integrated space-air-ground networks require mobile nodes to predict CSI in advance for optimizing trajectories and communication \cite{cheng2019space, wang2022joint}. Furthermore, 6G is evolving towards integrated sensing and communication (ISAC) architectures, leveraging existing radio frequency (RF) signals for environmental perception without dedicated sensing overhead \cite{zeng2024tutorial}. These demands collectively drive the development of environment-aware communication frameworks, where radio maps (RMs) serve as fundamental tools. RMs are expected to provide location-dependent, pilot-free CSI or channel features, enabling intelligent and adaptive 6G networks\cite{zeng2024tutorial}.

Despite their potential, constructing accurate RMs remains a significant challenge, especially in the context of next-generation 6G networks. While electromagnetic (EM) simulation methods such as ray tracing \cite{oh2004mimo} offer high precision by modeling wave propagation in detail, they are computationally prohibitive for large-scale and dynamic 6G environments. On the other hand, data-driven approaches, particularly those based on deep learning \cite{levie2021radiounet,10764739,10130091}, provide a scalable alternative by learning propagation patterns from data. However, these models heavily rely on large volumes of high-quality, diverse training data. Unfortunately, existing public datasets \cite{levie2021radiounet,10315088,10693754,10757328,10682510} are insufficient for comprehensive 6G research. They often lack the sub-meter spatial resolution required for advanced applications such as centimeter-level positioning, fine-grained beam management, and environment sensing. Moreover, most datasets focus only on basic metrics like path loss and are collected under single-input single-output (SISO) configurations, failing to represent the intricate channel dynamics of modern large-scale MIMO and beamformed systems operating at millimeter-wave and terahertz frequencies.

\begin{table*}[ht]
\vspace{5pt}
\centering 
\caption{Comparison of UrbanMIMOMap with Other Public Datasets.}
\label{tab:dataset_comparison}
\renewcommand{\arraystretch}{1.22}
\begin{adjustbox}{width=0.9\textwidth} 
\begin{tabular}{lccccc} 
\toprule
Dataset & Raw channel Matrix & Sub-meter accuracy & Large real scenario & MIMO support & Output map image \\ 
\midrule
UrbanMIMOMap(ours) & {\color{red!75!black} \ding{52}} & {\color{red!75!black} \ding{52}} & {\color{red!75!black} \ding{52}} & {\color{red!75!black} \ding{52}} & {\color{red!75!black} \ding{52}} \\ 
DeepMIMO \cite{alkhateeb2019deepmimo} & \ding{55} & \ding{52} & \ding{55} & \ding{52} & \ding{55} \\ 
CKMImageNet \cite{10693754} & \ding{55} & \ding{55} & \ding{52} & \ding{55} & \ding{52} \\ 
DataAI-6G \cite{10464657} & \ding{52} & \ding{52} & \ding{55} & \ding{52} & \ding{55} \\ 
M$^3$SC \cite{10315088} & \ding{52} & \ding{55} & \ding{52} & \ding{52} & \ding{55} \\ 
RadioMapSeer \cite{levie2021radiounet} & \ding{55} & \ding{55} & \ding{52} & \ding{55} & \ding{52} \\ 
SpectrumNet \cite{10757328} & \ding{55} & \ding{55} & \ding{52} & \ding{55} & \ding{52} \\ 
RadioGAT \cite{10682510} & \ding{55} & \ding{55} & \ding{52} & \ding{55} & \ding{52} \\ 
\bottomrule
\end{tabular}
\end{adjustbox}

\end{table*}

To address these limitations, we introduce UrbanMIMOMap, a novel large-scale urban MIMO RM dataset. This dataset captures realistic urban layouts and provides comprehensive complex CSI, overcoming the limitations of existing datasets. The rich CSI enables the derivation of various physical layer performance indicators and RMs, such as received signal strength (RSS) and theoretical channel capacity. Furthermore, the dataset supports data-driven RM generation, including deep learning methods, for which we provide baseline performance evaluations of representative neural network models.
The main contributions of this paper are summarized as follows.
\begin{enumerate}
    \item We present UrbanMIMOMap, a novel large-scale urban MIMO dataset generated via high-precision ray tracing. It features high spatial resolution (0.5m) and realistic environmental modeling, specifically designed to support advanced 6G wireless communication research.
    \item We provide complete complex MIMO channel matrices (H) per spatial point. This detailed information is crucial for advanced MIMO analysis and design, unlike datasets limited to simpler metrics.
    \item We demonstrate the dataset's practical utility by deriving key performance indicators from the raw CSI, such as precoding-aware RSS and theoretical channel capacity. Furthermore, we establish baseline performance benchmarks for representative machine learning models applied to RM construction tasks using this dataset.
\end{enumerate}

\section{The UrbanMIMOMap Dataset}
This section will first highlight the unique advantages and innovations of the UrbanMIMOMap dataset compared to existing public datasets. Subsequently, we will detail the generation process and methods employed for this dataset using specific examples.

\subsection{Key Features}
UrbanMIMOMap's core advantages (Table \ref{tab:dataset_comparison}) are summarized as follows.

\textbf{MIMO System Support.} UrbanMIMOMap provides a 4x4 MIMO configuration. Unlike SISO datasets offering a single channel coefficient ($h$), it delivers the complete channel matrix ($\mathbf{H}$). This matrix is crucial as it inherently captures the spatial correlation between antenna elements arising from coherent multipath effects. Such spatially correlated information cannot be accurately synthesized from separate SISO simulations, making our native MIMO data essential for realistically modeling multi-antenna systems in contrast to datasets like CKMImageNet \cite{10693754}.

\textbf{Complete Complex Channel Matrix.} UrbanMIMOMap provides the complete complex channel matrix ($\mathbf{H}$) at each point. In contrast to datasets limited to path loss, such as DeepMIMO \cite{alkhateeb2019deepmimo}, our matrix contains the precise amplitude and phase of every link. This detailed phase information, reflecting signal angles of arrival across the array, is indispensable for evaluating spatial processing like precoding, calculating true MIMO capacity, and leveraging the channel for sensing, making the dataset fundamentally more capable for advanced wireless research.

\textbf{Sub-meter Spatial Resolution.} The dataset features a 0.5m spatial resolution, crucial for 6G's fine-grained sensing and positioning needs. This is a significant improvement over datasets with coarser 1-5m resolutions, such as RadioMapSeer \cite{levie2021radiounet} and M$^3$SC \cite{10315088}. The resulting dense sampling provides a robust foundation for developing high-precision algorithms for tasks like mmWave beam management and fingerprint positioning.

\begin{figure}[htbp]
\captionsetup{font={small}, skip=8pt}
\centering 

\begin{tabular}{@{}ccc@{}} 
\includegraphics[width=0.26\linewidth]{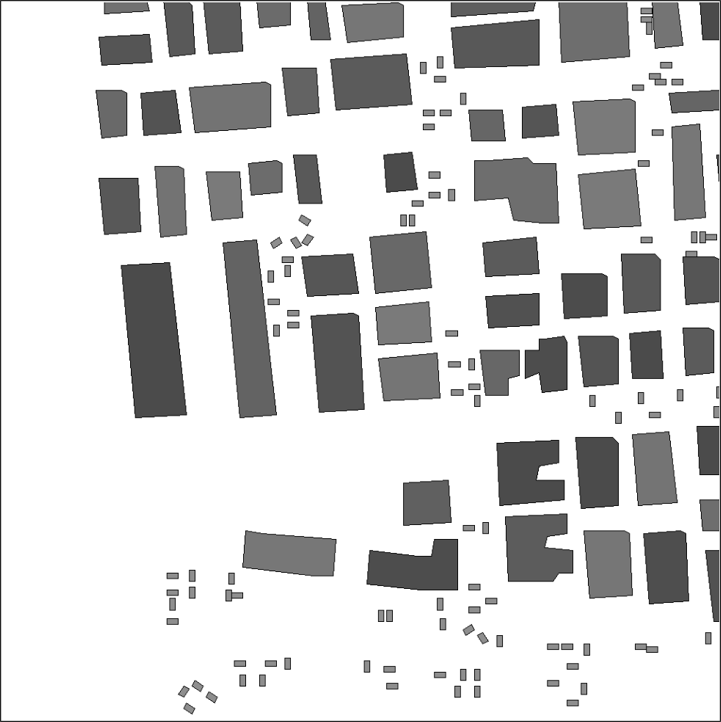} & 
\includegraphics[width=0.26\linewidth]{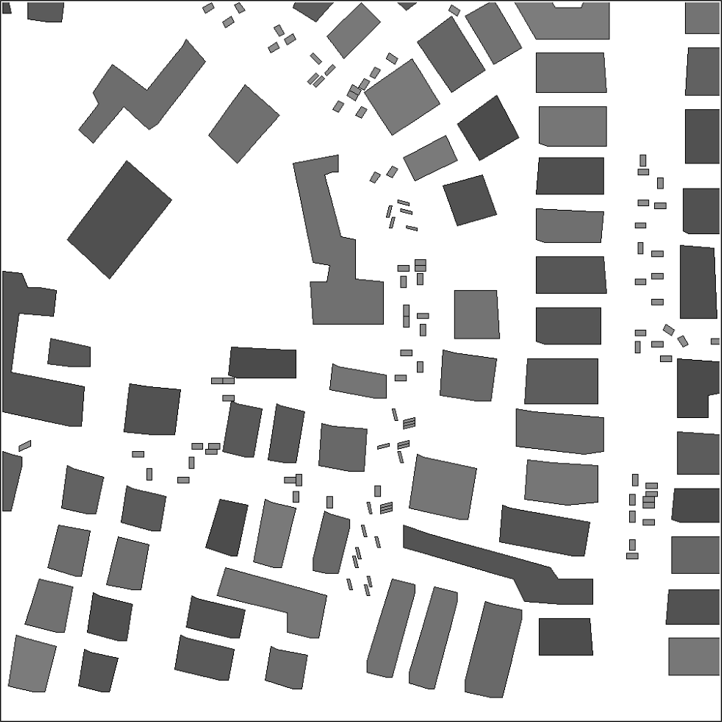} &
\includegraphics[width=0.26\linewidth]{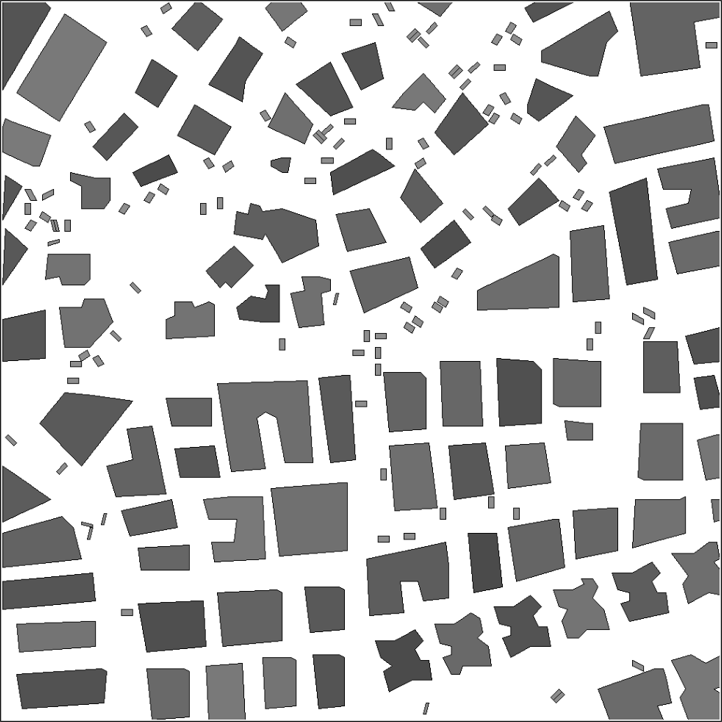} \\ 

\includegraphics[width=0.26\linewidth]{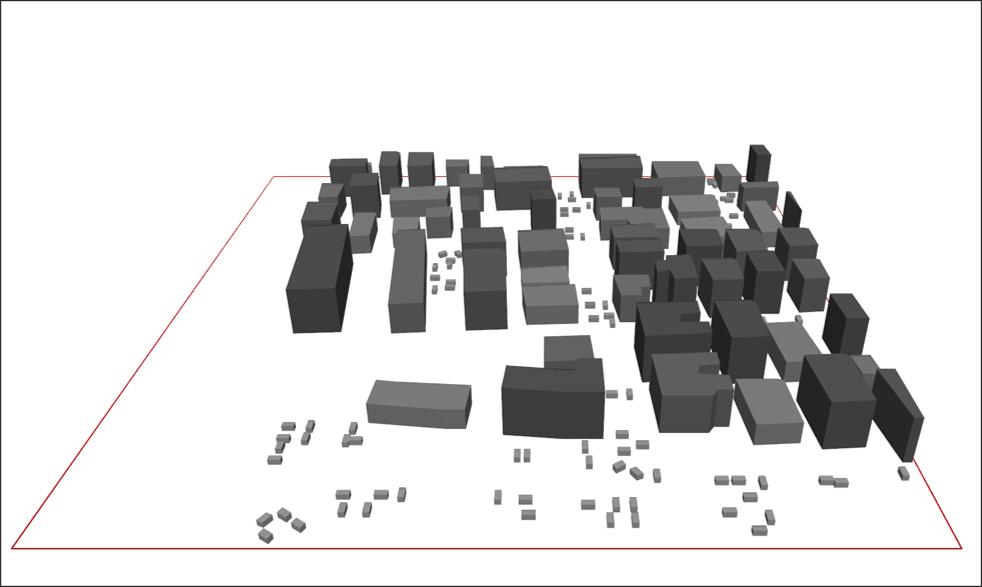} & 
\includegraphics[width=0.26\linewidth]{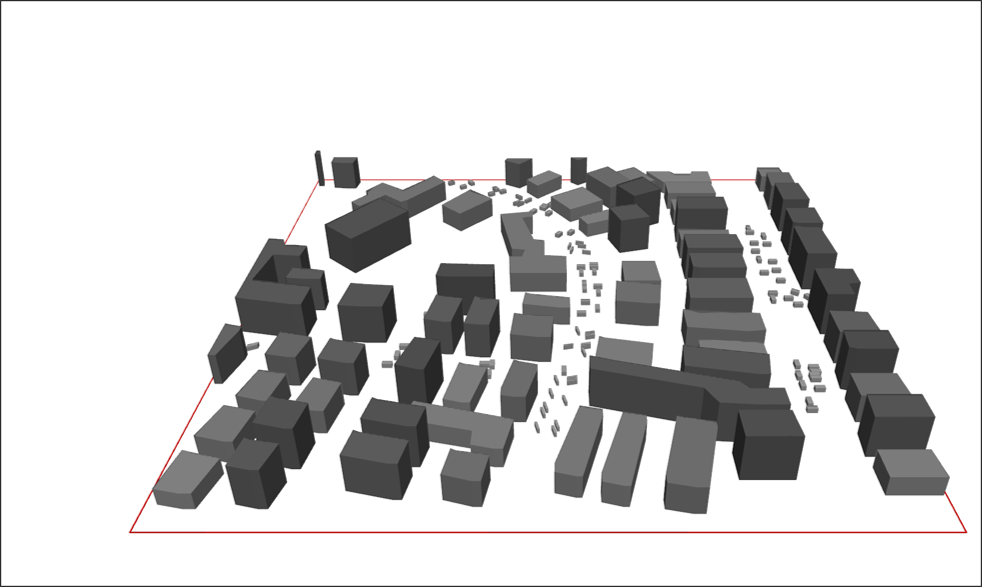} &
\includegraphics[width=0.26\linewidth]{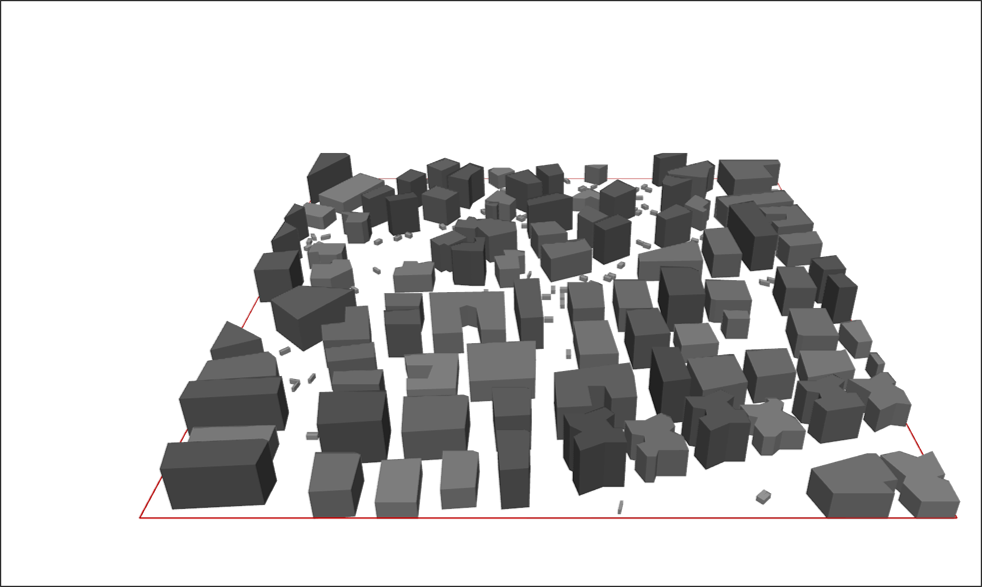} \\ 

\multicolumn{1}{c}{scene1} & 
\multicolumn{1}{c}{scene2} &
\multicolumn{1}{c}{scene3} \\ 

\end{tabular}

\caption{Example simulation scenes (top: 2D views, bottom: 3D views).}
\vspace{-9pt}
\label{fig:sim_scene}
\end{figure}

\textbf{Foundation on Realistic Urban Geometries.} Our dataset is built on large-scale, realistic urban geometries from OpenStreetMap, including detailed building and vehicle models (Fig.~\ref{fig:sim_scene}). This real-world complexity, in contrast to datasets based on simplified or smaller-scale scenarios like DeepMIMO \cite{alkhateeb2019deepmimo}, ensures that algorithms trained on UrbanMIMOMap are better suited for practical deployment.

\begin{figure*}[htbp]
\centering
\includegraphics[width=\linewidth]{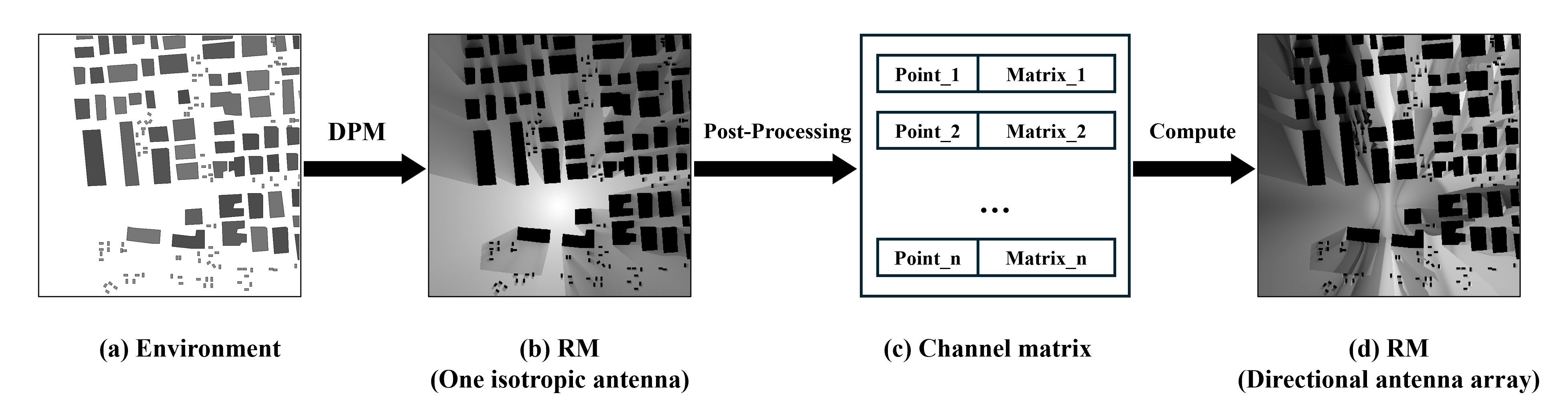} 
\caption{UrbanMIMOMap generation process overview.} 
\label{fig:gen_process}
\end{figure*}

\textbf{Visualizations and Data-Driven Support.} UrbanMIMOMap provides pre-rendered, visualizable radio maps. Unlike datasets that only offer raw channel data, such as DeepMIMO \cite{alkhateeb2019deepmimo} and M$^3$SC \cite{10315088}, these map images serve as direct inputs for image-based deep learning models. We demonstrate this by providing benchmarks for RM construction using the RadioUNet architecture \cite{levie2021radiounet}.

\textbf{Flexible Parameter Configuration.} The dataset's generation framework is highly flexible, allowing researchers to create custom datasets. Key parameters like MIMO array size, operating frequency, antenna patterns, and power levels can be easily modified. This configurability supports a broad range of investigations into how different system parameters impact channel and RM characteristics, valuable for future 6G research.

\subsection{Dataset Generation Process}
This section outlines the UrbanMIMOMap dataset generation process, detailing the simulation tools, core MIMO channel synthesis method, and key parameter settings with their justifications. Fig.~\ref{fig:gen_process} illustrates these key stages.

The UrbanMIMOMap dataset is generated through the following three main stages.

\textbf{Stage One: High-Precision Environment Modeling.}
High-precision 3D urban models are built from real geographic data (OpenStreetMap, e.g., Ankara, Berlin), including detailed buildings (6.6–19.8m) and vehicles (2m) for realism. Coordinates derive from processed RadioMapSeer data \cite{levie2021radiounet}. Fig.~\ref{fig:gen_process}(a) shows a top view (color indicates height). This modeling ensures representativeness, capturing urban multipath effects and supporting accurate channel simulation.

\textbf{Stage Two: Propagation Information.}
Propagation data is generated using Altair WinProp’s ray-tracing engine \cite{7916282} with the Dominant Path Model (DPM), balancing accuracy and speed for large urban areas. While Intelligent Ray Tracing captures more multipath details, it is computationally expensive. DPM enables fast generation by focusing on dominant paths, with slight simplification. An isotropic antenna at base stations emits signals; parameters are in Table~\ref{tab:sim_params_stage2}. WinProp computes path data (e.g., delay, DoD/DoA) for all 262,144 points—antenna-independent for later MIMO use. Fig.~\ref{fig:gen_process}(b) shows an example RSS map.

\begin{table}[htbp]
\centering
\caption{Simulation Parameters for Stage Two.} 
\label{tab:sim_params_stage2}
\begin{tabular}{llcr} 
\toprule
\multicolumn{2}{c}{\textbf{Parameter}} & \textbf{Value} & \textbf{Unit} \\ 
\midrule
\multirow{5}{*}{Simulation environment} & Map length & 256 & m \\
& Map width & 256 & m \\
& Prediction height & 1.5 & m \\
& Resolution & 0.5 & m \\
& Building height range & 6.6--19.8 & m \\ 
& Vehicle height & 2 & m \\ 
\midrule
\multirow{3}{*}{Transmitter} & Frequency & 3500 & MHz \\
& Bandwidth & 50 & MHz \\ 
& Power & 23 & dBm \\ 
& Antenna type & isotropic & $\sim$ \\ 
\bottomrule
\end{tabular}
\end{table}

\textbf{Stage Three: MIMO Channel Matrix Synthesis via Post-Processing.}
The final stage uses WinProp's MIMO post-processing to efficiently synthesize 4x4 complex channel matrices ($\mathbf{H}$). Unlike traditional methods requiring ray tracing for each antenna pair, this approach needs only one path simulation. Researchers can quickly evaluate various antenna parameters (patterns, configurations, orientations) on MIMO performance without repeated ray tracing, enhancing efficiency and flexibility. The process is as follows.
\begin{enumerate}
    \item Import the fine-grained propagation path information obtained in Stage Two, connecting the transmit location to all receive sampling points.
    \item In the post-processing step, define the actual MIMO transmit (Tx) and receive (Rx) antenna array configurations, including antenna element type, array geometry, element pattern, polarization, etc.
    \item Synthesize the complete MIMO channel matrix $\mathbf{H}$ by coherently summing the contributions of each path after passing through the corresponding antenna elements, based on the pre-stored path data and the characteristics of the MIMO antenna defined by the user.
\end{enumerate}
Detailed simulation parameters for this stage are provided in Table \ref{tab:antenna_params_stage3}. The key parameters are explained as follows.
\begin{itemize}
    \item \textbf{Inherited Parameters:} Parameters such as Tx or Rx antenna heights, carrier frequency, and transmit power are inherited from the Stage Two configuration.
    \item \textbf{Antenna Configuration:} Both the Tx and Rx ends are configured as 1-row, 4-column uniform linear arrays (ULA). The radiation pattern for each individual element within these arrays is defined by a specific beam pattern file. This configuration results in a 4x4 MIMO system. The framework supports using various antenna patterns and array configurations, including up to 64x64 MIMO.
    \item \textbf{Orientation and Array Parameters:} Key parameters include the overall orientation of the Tx ULA (azimuth 0°, 60°, 120°, tilt -10°) and the Rx ULA (azimuth 0°, tilt 90°). The spacing between elements in both ULAs is $\lambda/2$. Values for tilt, height, power, frequency, and bandwidth align with recommendations for urban deployments \cite{9390169}. Other parameters use WinProp's default values.
\end{itemize}

The dataset contains 262,144 sampling points, each providing a 4x4 complex MIMO channel matrix $\mathbf{H}$. Each $H_{ij}$ represents the complex channel gain from transmit antenna $j$ to receive antenna $i$, with its magnitude reflecting combined path loss, shadow fading, and small-scale multipath effects, and its phase encompassing total phase rotation from path length, reflection/diffraction shifts, and antenna array factors.

\begin{figure}[htbp]
\centering
\includegraphics[width=0.95\linewidth]{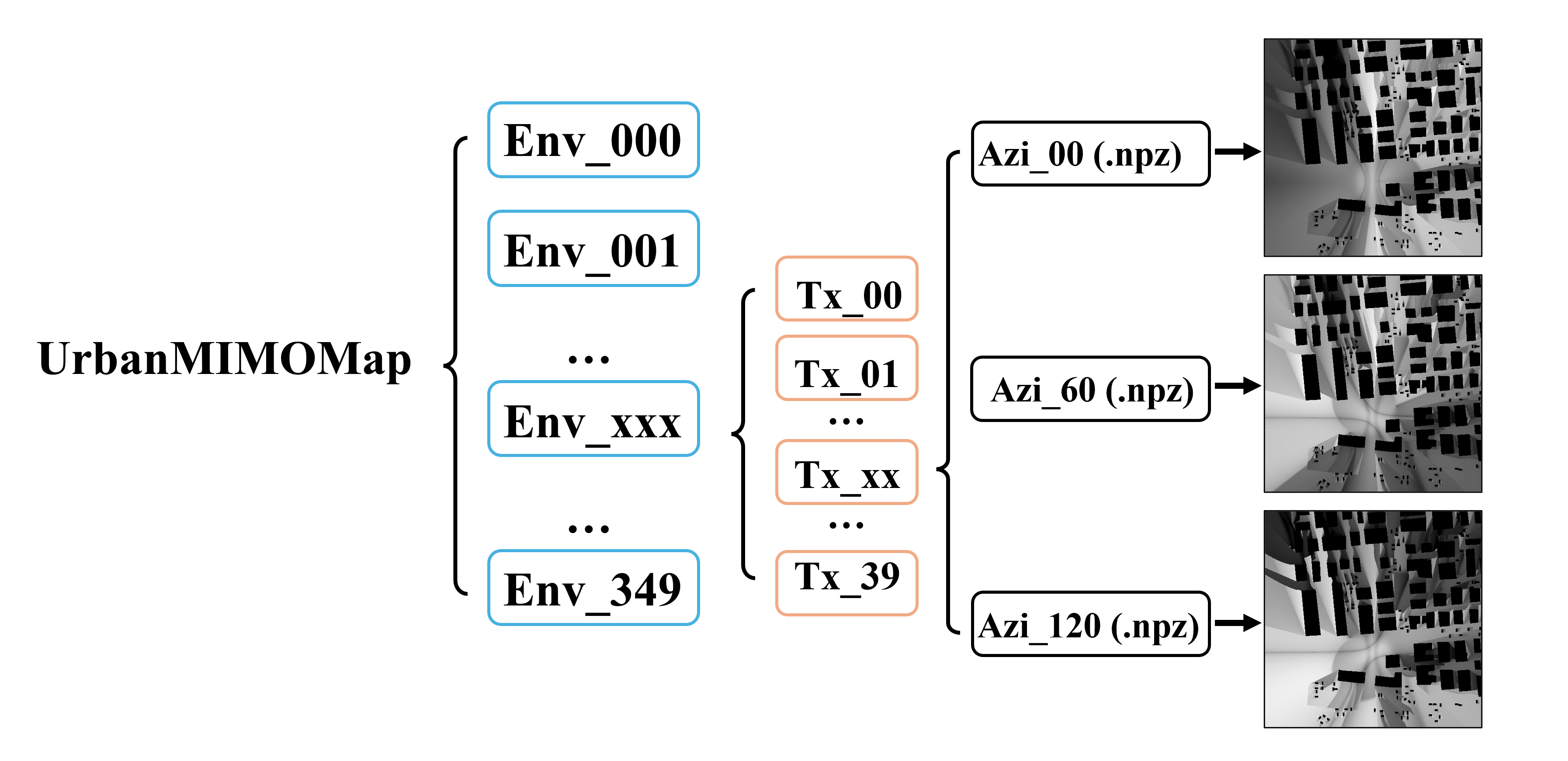} 
\caption{UrbanMIMOMap data structure and file organization.} 
\label{fig:dataset_arch}
\end{figure}

The dataset is organized into 350 scenarios with 40 transmitting base stations per scenario, and the antenna array at each base station is configured with three different azimuth angles, leading to a total of 42,000 numpy zipped files. Files are named by environment, transmitter, and azimuth, such as 0\_5\_60 and 0\_5\_120, with 0\_5\_iso indicating an isotropic antenna configuration. Fig.~\ref{fig:dataset_arch} illustrates this structure.

\begin{table*}[htbp]
\vspace*{5pt}
\centering
\caption{MIMO Configuration Parameters.} 
\label{tab:antenna_params_stage3}
\begin{tabular}{llccc} 
\toprule
\textbf{Parameter} & \textbf{Tx/Rx} & \textbf{Selected Value} & \textbf{Other Options} & \textbf{Unit} \\
\midrule
\multirow{2}{*}{Number of antennas} & Tx & 4 & 1 -- 64 & count \\ 
& Rx & 4 & 1 -- 64 & count \\
\midrule
\multirow{2}{*}{Arrangement type} & Tx & Linear (1x4 ULA) & Matrix, circular, etc. & $\sim$ \\ 
& Rx & Linear (1x4 ULA) & Matrix, circular, etc. & $\sim$ \\ 
\midrule
\multirow{2}{*}{Antenna type} & Tx & Beams2Control4Data.ffe & Custom patterns & $\sim$ \\
& Rx & Beams2Control4Data.ffe & Custom patterns & $\sim$ \\
\midrule
\multirow{2}{*}{Polarization} & Tx & Dual polarization & Vertical, circular, etc. & $\sim$ \\
& Rx & Dual polarization & Vertical, circular, etc. & $\sim$ \\
\midrule
\multirow{2}{*}{Azimuth angle} & Tx & 0, 60, 120 & 0 -- 360 & degrees \\
& Rx & 0 & 0 -- 360 & degrees \\ 
\midrule
\multirow{2}{*}{Elevation tilt} & Tx & -10 & -90 -- 90 & degrees \\ 
& Rx & 90 & -90 -- 90 & degrees \\ 
\midrule
\multirow{2}{*}{Antenna spacing} & Tx & 0.5 & $\sim$ & Wavelength multiple \\ 
& Rx & 0.5 & $\sim$ & Wavelength multiple \\ 
\midrule
\multirow{2}{*}{Coupling} & Tx & False & True & $\sim$ \\
& Rx & False & True & $\sim$ \\
\midrule
\multicolumn{2}{l}{Coherent signal addition} & True & False & $\sim$ \\ 
\bottomrule
\end{tabular}
\end{table*}

\section{Application Examples}
UrbanMIMOMap's complete MIMO CSI enables diverse research applications. This section presents key examples demonstrating its utility.

\textbf{Case 1: RM Generation and Analysis under Specific Precoding Strategies.}

A key advantage of UrbanMIMOMap is its complete MIMO channel matrix $\mathbf{H}$, enabling generation and analysis of RMs under various transmission strategies. As an illustration, we consider fixed equal gain transmission (FEGT) precoding. While FEGT is simple yet effective, the dataset's rich CSI allows researchers to evaluate more sophisticated precoding and beamforming strategies, such as zero-forcing (ZF), by leveraging the provided H matrices in post-processing. The FEGT precoding vector is as follows.
\begin{equation}
\mathbf{w} = \frac{1}{2} \begin{bmatrix}1 & 1 & 1 & 1\end{bmatrix}^T, \label{eq:egt_w} 
\end{equation}
where the coefficient $1/2$ ensures power normalization ($\|\mathbf{w}\|^2=1$). Under a single-symbol transmission assumption with symbol $s$ satisfying $|s|^2=1$, the transmitted signal is $\mathbf{x} = \mathbf{w}s$. The received signal vector $\mathbf{y}$ is as follows.
\begin{equation}
\mathbf{y} = \mathbf{H}\mathbf{w}s + \mathbf{n},
\end{equation}
where $\mathbf{n}$ is the noise vector. The received signal power $P_{\text{rx}}$ at each location (neglecting noise for this calculation) is determined by the squared norm of the effective channel vector, $\mathbf{H}\mathbf{w}$. $P_{\text{rx}}$ can be represented as follows.
\begin{equation}
P_{\text{rx}} = \left\|\mathbf{H} \mathbf{w}\right\|^2 \|\mathbf{s}\|^2 = \left\|\mathbf{H} \mathbf{w}\right\|^2,
\end{equation}
Converting to dBm units as follows.
\begin{equation}
P_{\text{rx, dBm}} = 10 \log_{10}(P_{\text{rx}}) + 30, 
\end{equation}
The resulting RSS values in dBm are then linearly normalized to the [0, 1] range for visualization as follows.
\begin{equation}
P_{\text{rx, normalized}} = \frac{P_{\text{rx, dBm}} - P_{\min, \text{dBm}}}{P_{\max, \text{dBm}} - P_{\min, \text{dBm}}}.
\label{eq:rss_normalization}
\end{equation}
where $P_{\min, \text{dBm}}$ and $P_{\max, \text{dBm}}$ are the minimum and maximum RSS values in dBm units across the whole dataset.

\begin{figure*}[htbp]
\captionsetup{font={small}, skip=2pt}
\centering 
\begin{tabular}{@{}cccc@{}} %
\includegraphics[width=0.2\linewidth]{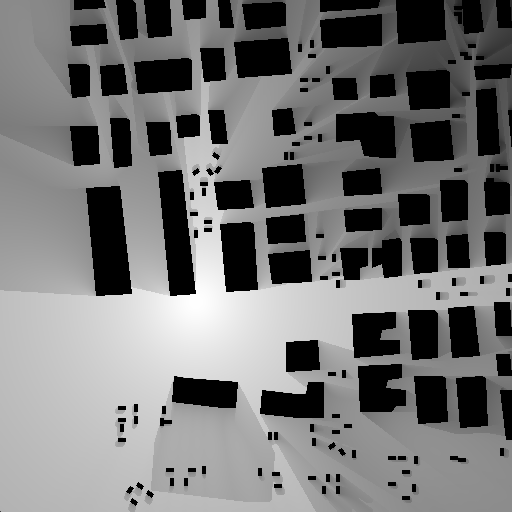} &
\includegraphics[width=0.2\linewidth]{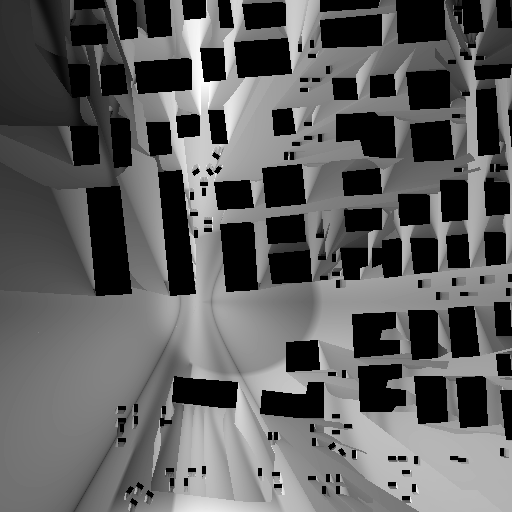} &
\includegraphics[width=0.2\linewidth]{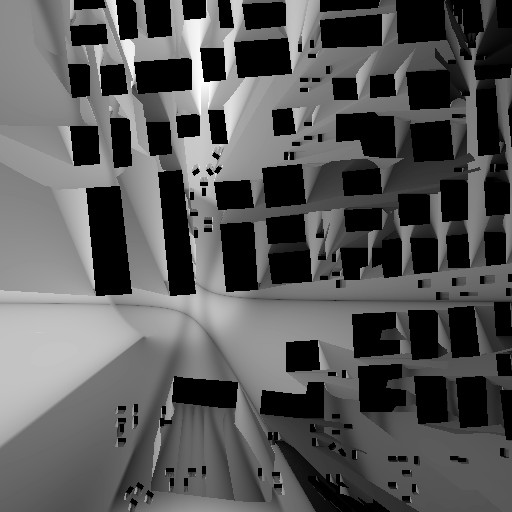} &
\includegraphics[width=0.2\linewidth]{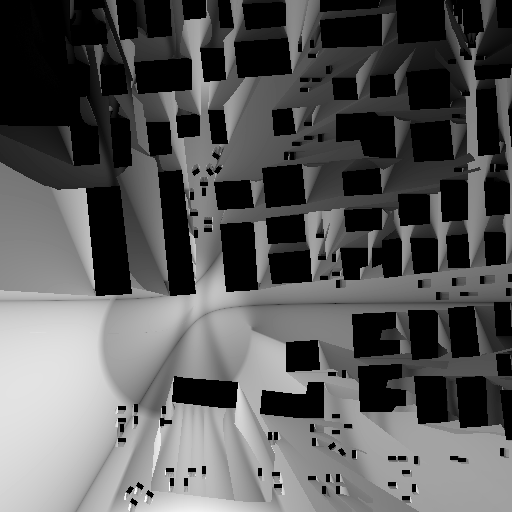} \\ 
\multicolumn{1}{c}{0\_5\_iso} & 
\multicolumn{1}{c}{0\_5\_0} &
\multicolumn{1}{c}{0\_5\_60} &
\multicolumn{1}{c}{0\_5\_120} \\
\end{tabular}
\caption{Comparison: isotropic path loss RM vs. FEGT-RSS RM from UrbanMIMOMap.}
\vspace{-9pt}
\label{fig:rm_comparison}
\end{figure*}

Fig.~\ref{fig:rm_comparison} compares the generated FEGT-RSS maps with isotropic path loss maps, illustrating the impact of antenna array's directivity. The dataset's $\mathbf{H}$ matrices allow evaluating various precoding or beamforming strategies.

To demonstrate UrbanMIMOMap's applicability in data-driven RM generation using machine learning, we employed the RadioUNet architecture \cite{levie2021radiounet} for a RM prediction task. For this baseline, we designed a multi-channel input tensor $\mathbf{I}$ to provide the network with essential environmental and configurational context, aiding reproducibility and further development. The input is a 4-channel tensor of 512x512 size, with channels $\mathbf{I}_c$ composed as follows:
\begin{itemize}
    \item Channel 1 ($\mathbf{I}_1$): Upsampled environment map (from 256$\times$256 to 512$\times$512 pixels using nearest-neighbor interpolation), vehicle information is already included.
    \item Channel 2 ($\mathbf{I}_2$): Binary map indicating transmitter locations, derived by upscaling the 256x256 binary image to 512x512 pixels. The original 256x256 image encodes transmitter locations, with $\mathbf{I}_2(i, j) = 1$ at scaled transmitter pixels, and 0 otherwise.
    \item Channels 3 ($\mathbf{I}_3$) and 4 ($\mathbf{I}_4$): Encode the antenna azimuth angle $\theta_{azi}$ at transmitter locations, and are zero elsewhere. The encoding for a pixel $(i, j)$ is as follows.
    \begin{align}
    \mathbf{I}_3(i, j) &= \begin{cases} \sin(\theta_{azi}) & \text{if } \mathbf{I}_2(i, j) = 1 \\ 0 & \text{if } \mathbf{I}_2(i, j) = 0 \end{cases}, \\
    \mathbf{I}_4(i, j) &= \begin{cases} \cos(\theta_{azi}) & \text{if } \mathbf{I}_2(i, j) = 1 \\ 0 & \text{if } \mathbf{I}_2(i, j) = 0 \end{cases}.
    \end{align}
\end{itemize}

\begin{figure*}[t]
\captionsetup{font={small}, skip=2pt}
\centering 

\begin{tabular}{@{}cccccc@{}} %
\includegraphics[width=0.13\linewidth]{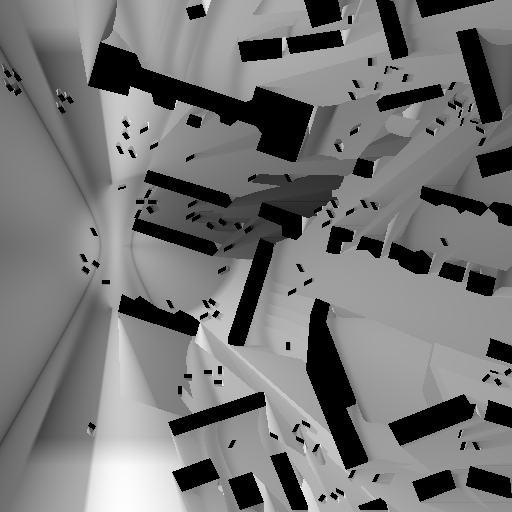} &
\includegraphics[width=0.13\linewidth]{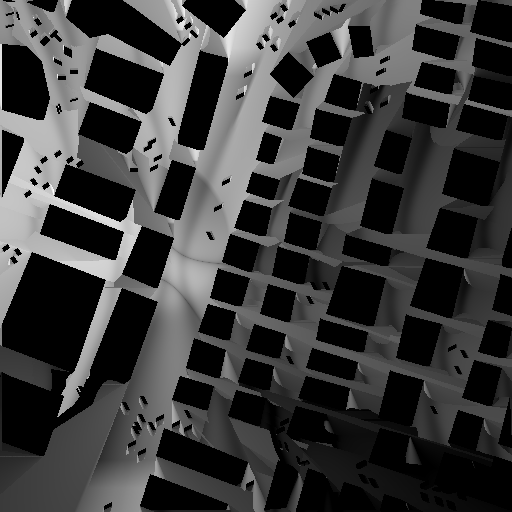} &
\includegraphics[width=0.13\linewidth]{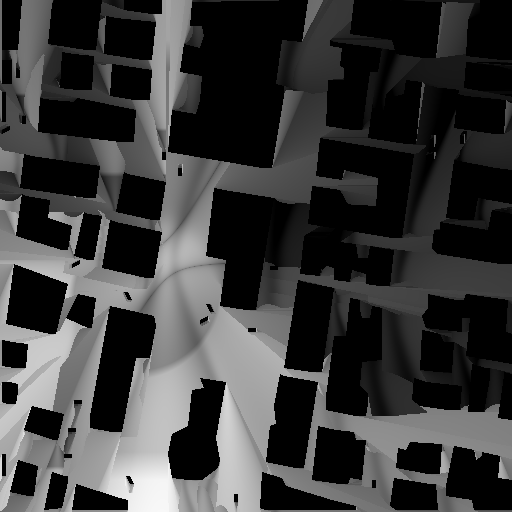} &
\includegraphics[width=0.13\linewidth]{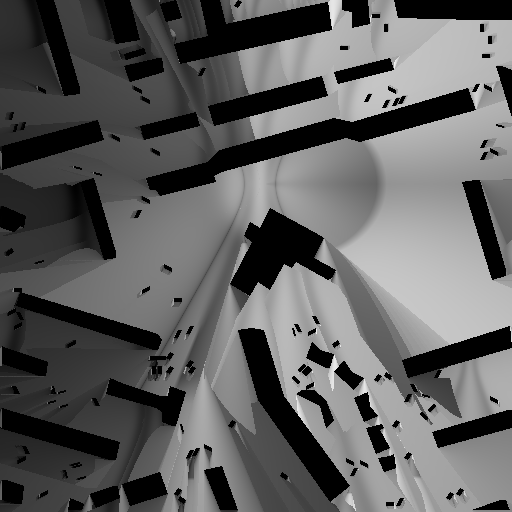} &
\includegraphics[width=0.13\linewidth]{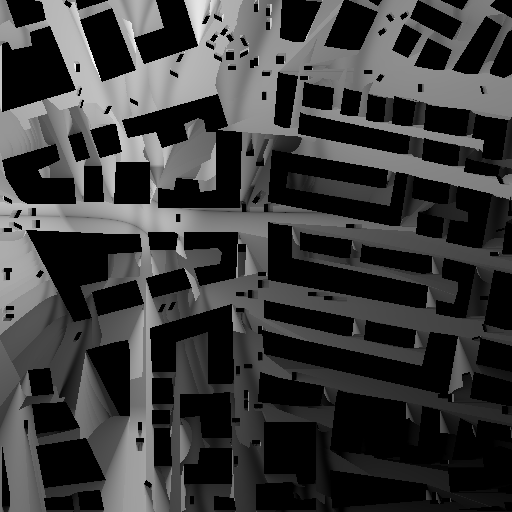} &
\includegraphics[width=0.13\linewidth]{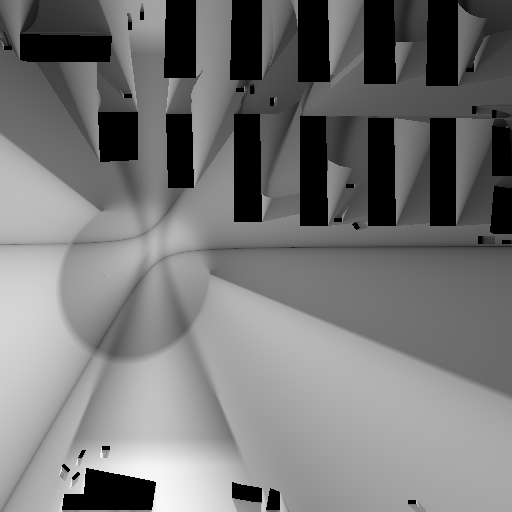} \\

\multicolumn{1}{c}{GT\_Azi\_0} &
\multicolumn{1}{c}{GT\_Azi\_60} &
\multicolumn{1}{c}{GT\_Azi\_120} &
\multicolumn{1}{c}{GT\_Azi\_0} &
\multicolumn{1}{c}{GT\_Azi\_60} &
\multicolumn{1}{c}{GT\_Azi\_120} \\

\includegraphics[width=0.13\linewidth]{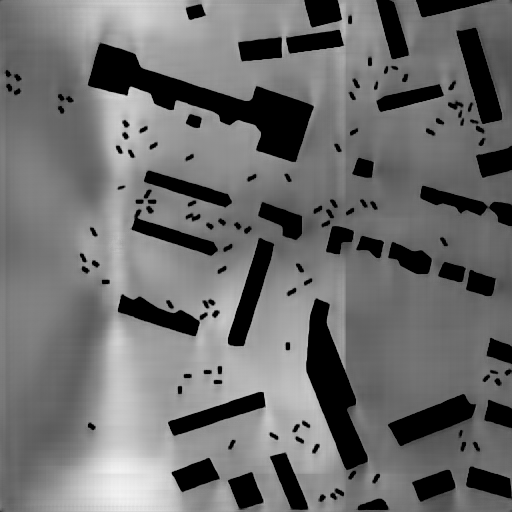} &
\includegraphics[width=0.13\linewidth]{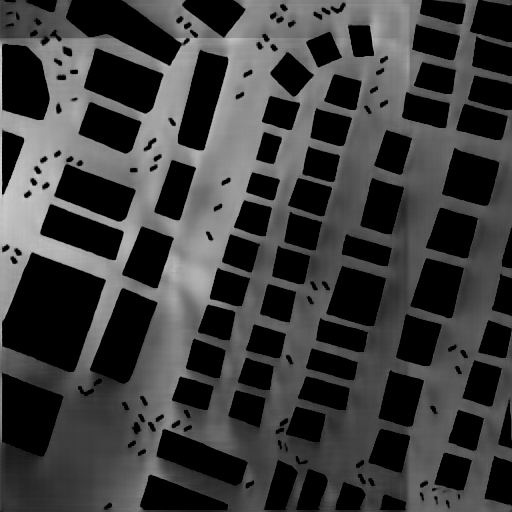} &
\includegraphics[width=0.13\linewidth]{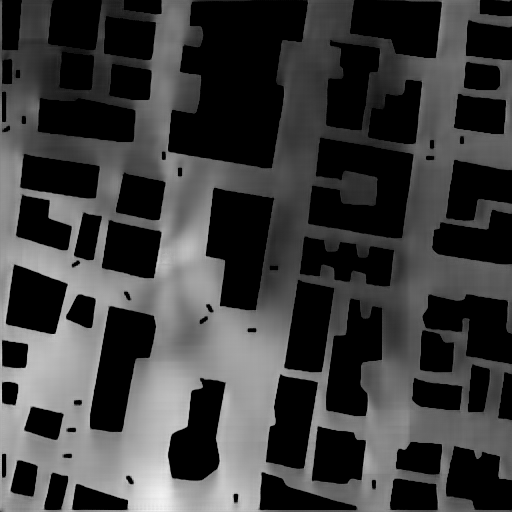} &
\includegraphics[width=0.13\linewidth]{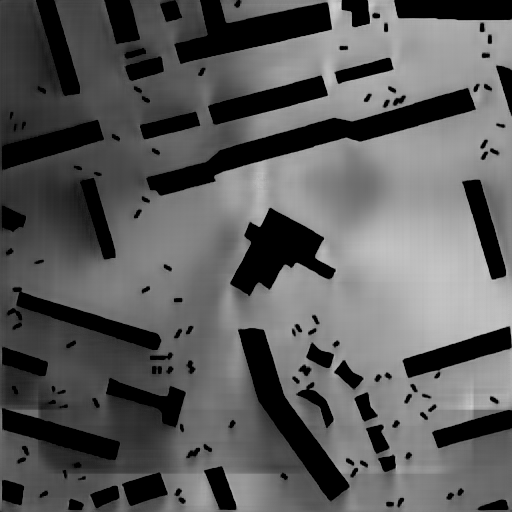} &
\includegraphics[width=0.13\linewidth]{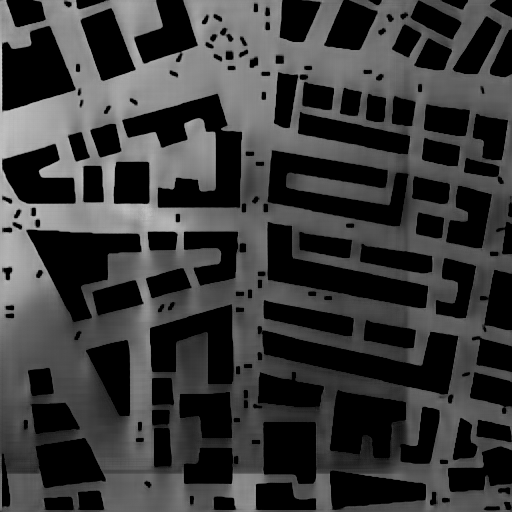} &
\includegraphics[width=0.13\linewidth]{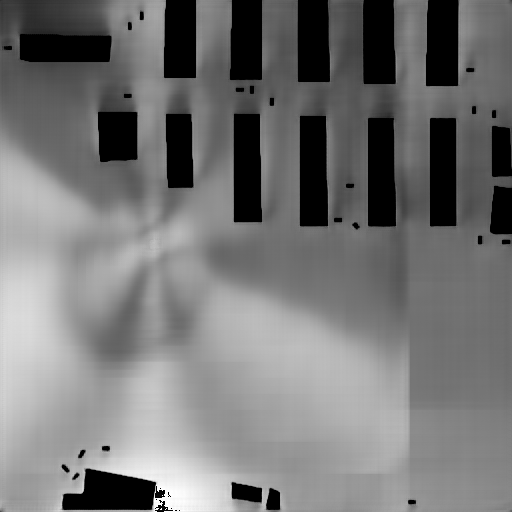} \\ 

\multicolumn{1}{c}{Pred\_Azi\_0} &
\multicolumn{1}{c}{Pred\_Azi\_60} &
\multicolumn{1}{c}{Pred\_Azi\_120} &
\multicolumn{1}{c}{Pred\_Azi\_0} &
\multicolumn{1}{c}{Pred\_Azi\_60} &
\multicolumn{1}{c}{Pred\_Azi\_120} \\ 

\end{tabular}

\caption{Examples of RadioUNet predictions for FEGT-RSS maps. The top row shows the ground truth maps, and the bottom row displays the corresponding predicted maps generated by the model.}
\vspace{-9pt}
\label{fig:radiounet_results}
\end{figure*}

We trained the baseline model for 30 epochs using the dataset, employing a split of 30,000 images for training, 6,000 for validation, and 6,000 for testing. We then evaluated its prediction performance. Quantitative results on the test set, including metrics such as normalized mean squared error (NMSE), root mean squared error (RMSE), structural similarity index measure (SSIM), and peak signal-to-noise ratio (PSNR), are presented in Table \ref{tab:radiounet_results}, while representative visual predictions are shown in Fig.~\ref{fig:radiounet_results}.

\begin{table}[htbp]
\centering
\caption{RadioUNet Baseline Performance for FEGT-RSS Prediction}
\label{tab:radiounet_results}
\small 
\renewcommand{\arraystretch}{1.22}

\begin{tabular}{cccc}
\toprule
NMSE & RMSE & SSIM & PSNR \\
\midrule
0.0793 & 0.1225 & 0.7989 & 18.2470 \\
\bottomrule
\end{tabular}
\vspace{-6pt} 
\end{table}

\begin{figure*}[htbp]
\captionsetup{font={small}, skip=2pt}
\centering 

\begin{tabular}{@{}cccccc@{}} %
\includegraphics[width=0.13\linewidth]{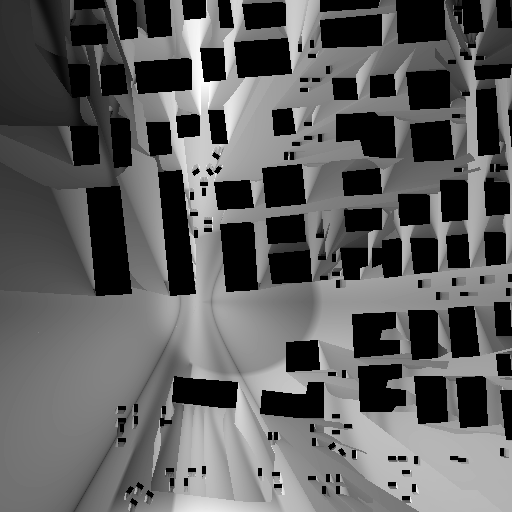} &
\includegraphics[width=0.13\linewidth]{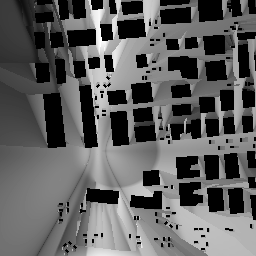} &
\includegraphics[width=0.13\linewidth]{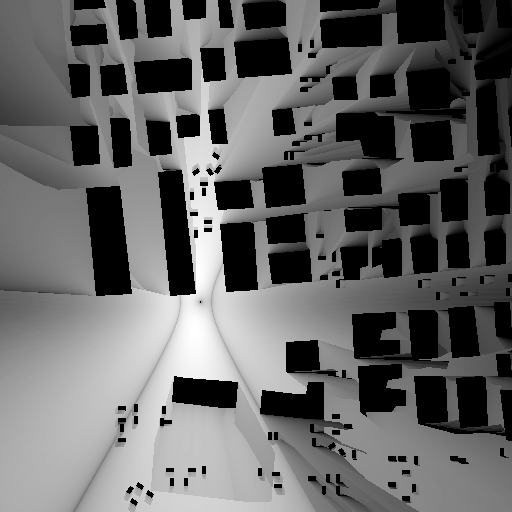} &
\includegraphics[width=0.13\linewidth]{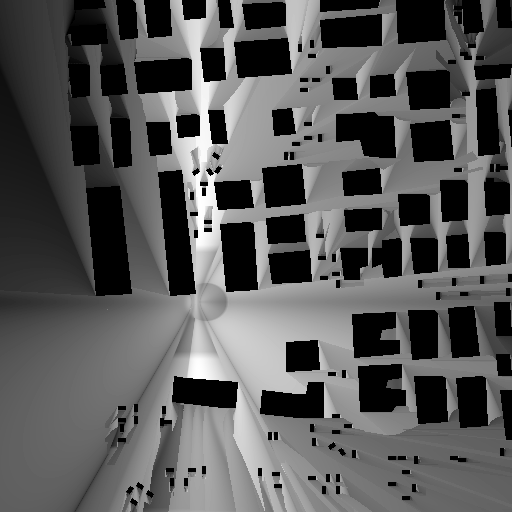} &
\includegraphics[width=0.13\linewidth]{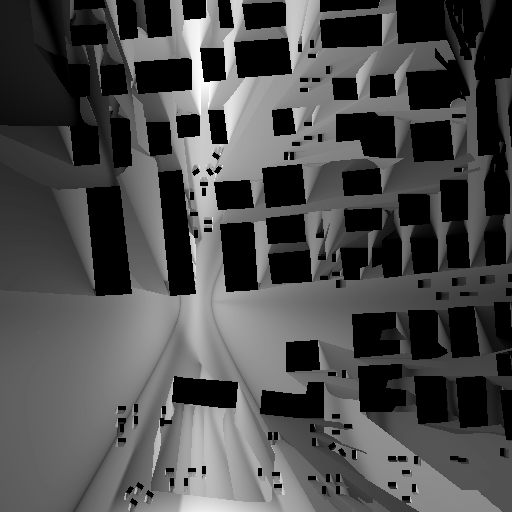} &
\includegraphics[width=0.13\linewidth]{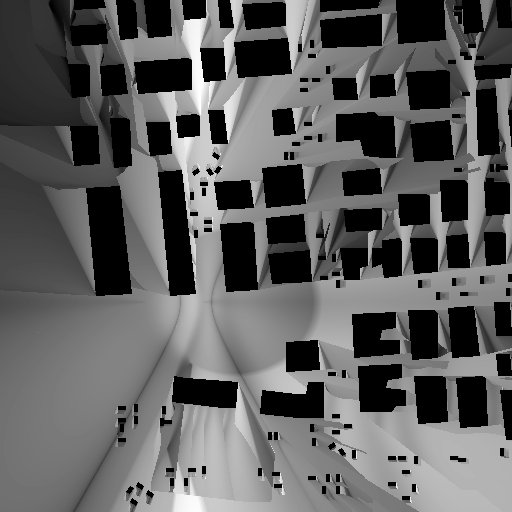} \\

\multicolumn{1}{c}{0\_5\_0} &
\multicolumn{1}{c}{Resolution:1.0m} &
\multicolumn{1}{c}{Pattern:dipole} &
\multicolumn{1}{c}{Tx\_height:5m} &
\multicolumn{1}{c}{Tx\_tilt:-30°} &
\multicolumn{1}{c}{Rx\_tilt:60°} \\
\end{tabular}

\caption{Illustrative examples showing how varying simulation parameters affect the generated RM. These are examples only and are not included in the dataset release.}
\vspace{-9pt}
\label{fig:param_comparison}
\end{figure*}

UrbanMIMOMap's suitability for training deep learning RM construction models is validated by this benchmark. The results from this basic implementation serve as an initial reference, pointing towards future research directions such as optimizing input representations (e.g., azimuth encoding) and network architectures for enhanced MIMO performance.

Furthermore, the generation framework allows creating custom datasets to study the impact of parameters like antenna height, tilt, pattern, and resolution on RMs (see Fig.~\ref{fig:param_comparison} for illustrative examples).

\textbf{Case 2: MIMO Channel Capacity Calculation and Analysis using CSI Data.}

Channel capacity, a fundamental metric in information theory, quantifies the theoretical maximum error-free transmission rate. The complete MIMO channel matrix $\mathbf{H}$ provided by UrbanMIMOMap enables the calculation of theoretical ergodic channel capacity at each location. Analyzing the spatial distribution of capacity provides insights into potential MIMO performance in urban areas, offering references for network planning, resource allocation, and optimization.

Assuming Gaussian inputs, equal power distribution across $N_t$ transmit antennas, and $\mathbf{H}$ known at the receiver, the theoretical ergodic channel capacity $C$ (in bits/s/Hz) is calculated as follows.
\begin{equation}
C = \log_2 \det\left(\mathbf{I}_{N_r} + \frac{\rho}{N_t} \mathbf{H} \mathbf{H}^H\right),
\label{eq:mimo_capacity} 
\end{equation}
where $\mathbf{I}_{N_r}$ is the $N_r \times N_r$ identity matrix, $N_t=4$, $N_r=4$, $\mathbf{H}^H$ is the conjugate transpose (Hermitian transpose) of $\mathbf{H}$, and $\rho$ is the average SNR at each receiver antenna element. Applying this formula across all sampling points allows for the calculation and visualization of channel capacity maps, analogous to the RSS maps shown in Case 1.

\section{Conclusion}
In this paper, we have introduced UrbanMIMOMap, a novel, large-scale urban MIMO RM dataset generated using high-precision ray tracing within realistic city models. UrbanMIMOMap provides complete complex CSI matrices at a high spatial resolution (0.5m). This rich dataset enables detailed MIMO channel characterization, evaluation of spatial performance metrics, and serves as a basis for developing and benchmarking data-driven RM generation. UrbanMIMOMap is a vital resource for environment-aware paradigms like ISAC requiring fine-grained spatial channel knowledge. Furthermore, while the use of specific urban geometries may influence the generalizability of trained models to different city layouts, the flexible generation framework presented offers the potential for creating customized datasets tailored to broader research inquiries, helping to address this limitation.

\bibliography{ref}

\begin{thebibliography}{10}
\providecommand{\url}[1]{#1}
\csname url@samestyle\endcsname
\providecommand{\newblock}{\relax}
\providecommand{\bibinfo}[2]{#2}
\providecommand{\BIBentrySTDinterwordspacing}{\spaceskip=0pt\relax}
\providecommand{\BIBentryALTinterwordstretchfactor}{4}
\providecommand{\BIBentryALTinterwordspacing}{\spaceskip=\fontdimen2\font plus
\BIBentryALTinterwordstretchfactor\fontdimen3\font minus \fontdimen4\font\relax}
\providecommand{\BIBforeignlanguage}[2]{{%
\expandafter\ifx\csname l@#1\endcsname\relax
\typeout{** WARNING: IEEEtran.bst: No hyphenation pattern has been}%
\typeout{** loaded for the language `#1'. Using the pattern for}%
\typeout{** the default language instead.}%
\else
\language=\csname l@#1\endcsname
\fi
#2}}
\providecommand{\BIBdecl}{\relax}
\BIBdecl

\bibitem{bjornson2017massive}
E.~Bj{\"o}rnson, J.~Hoydis, L.~Sanguinetti \emph{et~al.}, ``{Massive MIMO networks}: Spectral, energy, and hardware efficiency,'' \emph{Foundations and Trends{\textregistered} in Signal Processing}, vol.~11, no. 3-4, pp. 154--655, 2017.

\bibitem{sun2025acomprehesive}
R.~Sun, N.~Cheng, C.~Li, W.~Quan, H.~Zhou, Y.~Wang, W.~Zhang, and S.~X. (Sherman), ``A comprehensive survey of knowledge-driven deep learning for intelligent wireless network optimization in 6g,'' \emph{IEEE Communications Surveys \& Tutorials}, 2025, early access.

\bibitem{wang2024tutorial}
Z.~Wang, J.~Zhang, H.~Du, D.~Niyato, S.~Cui, B.~Ai, M.~Debbah, K.~B. Letaief, and H.~V. Poor, ``A tutorial on extremely large-scale {MIMO} for {6G}: Fundamentals, signal processing, and applications,'' \emph{{IEEE} Commun. Surveys Tuts.}, vol.~26, no.~3, pp. 1560--1605, 2024.

\bibitem{cheng2019space}
N.~Cheng, F.~Lyu, W.~Quan, C.~Zhou, H.~He, W.~Shi, and X.~Shen, ``Space/aerial-assisted computing offloading for {IoT} applications: A learning-based approach,'' \emph{{IEEE} J. Select. Areas Commun.}, vol.~37, no.~5, pp. 1117--1129, 2019.

\bibitem{wang2022joint}
X.~Wang, L.~Fu, N.~Cheng, R.~Sun, T.~Luan, W.~Quan, and K.~Aldubaikhy, ``Joint flying relay location and routing optimization for {6G} {UAV}--{IoT} networks: A graph neural network-based approach,'' \emph{Remote Sens.}, vol.~14, no.~17, p. 4377, 2022.

\bibitem{zeng2024tutorial}
Y.~Zeng, J.~Chen, J.~Xu, D.~Wu, X.~Xu, S.~Jin, X.~Gao, D.~Gesbert, S.~Cui, and R.~Zhang, ``A tutorial on environment-aware communications via channel knowledge map for {6G},'' \emph{{IEEE} Commun. Surveys Tuts.}, vol.~26, no.~3, pp. 1478--1519, 2024.

\bibitem{oh2004mimo}
S.-H. Oh and N.-H. Myung, ``{MIMO} channel estimation method using ray-tracing propagation model,'' \emph{Electronics letters}, vol.~40, no.~21, pp. 1350--1352, 2004.

\bibitem{levie2021radiounet}
R.~Levie, {\c{C}}.~Yapar, G.~Kutyniok, and G.~Caire, ``{RadioUNet}: Fast radio map estimation with convolutional neural networks,'' \emph{IEEE Transactions on Wireless Communications}, vol.~20, no.~6, pp. 4001--4015, 2021.

\bibitem{10764739}
X.~Wang, K.~Tao, N.~Cheng, Z.~Yin, Z.~Li, Y.~Zhang, and X.~Shen, ``{RadioDiff}: An effective generative diffusion model for sampling-free dynamic radio map construction,'' \emph{IEEE Transactions on Cognitive Communications and Networking}, vol.~11, no.~2, pp. 738--750, 2025.

\bibitem{10130091}
S.~Zhang, A.~Wijesinghe, and Z.~Ding, ``{RME-GAN}: A learning framework for radio map estimation based on conditional generative adversarial network,'' \emph{IEEE Internet of Things Journal}, vol.~10, no.~20, pp. 18\,016--18\,027, 2023.

\bibitem{10315088}
X.~Cheng, Z.~Huang, L.~Bai, H.~Zhang, M.~Sun, B.~Liu, S.~Li, J.~Zhang, and M.~Lee, ``{M$^3$SC}: A generic dataset for mixed multi-modal (mmm) sensing and communication integration,'' \emph{China Communications}, vol.~20, no.~11, pp. 13--29, 2023.

\bibitem{10693754}
D.~Wu, Z.~Wu, Y.~Qiu, S.~Fu, and Y.~Zeng, ``{CKMI}magenet: A comprehensive dataset to enable channel knowledge map construction via computer vision,'' in \emph{2024 IEEE/CIC International Conference on Communications in China (ICCC Workshops)}, 2024, pp. 114--119.

\bibitem{10757328}
S.~Zhang, S.~Jiang, W.~Lin, Z.~Fang, K.~Liu, H.~Zhang, and K.~Chen, ``{Generative AI on SpectrumNet}: An open benchmark of multiband {3-D} radio maps,'' \emph{IEEE Transactions on Cognitive Communications and Networking}, vol.~11, no.~2, pp. 886--901, 2025.

\bibitem{10682510}
X.~Li, S.~Zhang, H.~Li, X.~Li, L.~Xu, H.~Xu, H.~Mei, G.~Zhu, N.~Qi, and M.~Xiao, ``{RadioGAT}: A joint model-based and data-driven framework for multi-band radiomap reconstruction via graph attention networks,'' \emph{IEEE Transactions on Wireless Communications}, vol.~23, no.~11, pp. 17\,777--17\,792, 2024.

\bibitem{alkhateeb2019deepmimo}
A.~Alkhateeb, ``{DeepMIMO}: A generic deep learning dataset for millimeter wave and massive {MIMO} applications,'' \emph{arXiv preprint arXiv:1902.06435}, 2019.

\bibitem{10464657}
Z.~Shen, L.~Yu, Y.~Zhang, J.~Zhang, Z.~Zhang, X.~Hu, S.~Han, J.~Jin, and G.~Liu, ``{DataAI--6G}: A system parameters configurable channel dataset for {AI-6G} research,'' in \emph{2023 IEEE Globecom Workshops (GC Wkshps)}, 2023, pp. 1910--1915.

\bibitem{7916282}
R.~Hoppe, G.~Wölfle, and U.~Jakobus, ``Wave propagation and radio network planning software {WinProp} added to the electromagnetic solver package {FEKO},'' in \emph{2017 International Applied Computational Electromagnetics Society Symposium - Italy (ACES)}, 2017, pp. 1--2.

\bibitem{9390169}
H.~Tataria, M.~Shafi, A.~F. Molisch, M.~Dohler, H.~Sjöland, and F.~Tufvesson, ``{6G} wireless systems: Vision, requirements, challenges, insights, and opportunities,'' \emph{Proceedings of the IEEE}, vol. 109, no.~7, pp. 1166--1199, 2021.

\end{thebibliography}
\bibliographystyle{IEEEtran}

\ifCLASSOPTIONcaptionsoff
  \newpage
\fi

\end{document}